%% file: main.tex
\documentclass[pdflatex,sn-basic]{sn-jnl}

\usepackage{graphicx}
\usepackage{amsmath,amssymb,amsfonts}
\usepackage{amsthm}
\usepackage{booktabs}
\usepackage{tabularx}
\usepackage{array}
\usepackage{microtype}
\usepackage{xurl}
\hypersetup{
  pdftitle={Checkpoint Selection and Evaluation in EEG Emotion Recognition},
  pdfauthor={Hanting Suo; Hongxun Wang; Yuwen Li},
  pdfsubject={Evaluation protocols for EEG emotion recognition},
  pdfkeywords={affective computing; benchmark reproducibility; test-set reuse; model selection; checkpoint search; subject-level inference}
}
\providecommand{\doi}[1]{%
  \ifstrequal{#1}{10.1186/1471-2105-7-91}{%
    \hfill\break\href{https://doi.org/#1}{\nolinkurl{https://doi.org/10.1186/1471-2105-7-91}}%
  }{\url{https://doi.org/#1}}%
}
\newcolumntype{Y}{>{\raggedright\arraybackslash}X}
\theoremstyle{plain}

\theoremstyle{definition}

\begin{document}

\title[Evaluation protocols for EEG emotion recognition]{Checkpoint Selection and Evaluation in EEG Emotion Recognition}

\author[1,2]{\fnm{Hanting} \sur{Suo}}
\author[1,2]{\fnm{Hongxun} \sur{Wang}}
\author*[1,2]{\fnm{Yuwen} \sur{Li}}\email{liyuwen@seu.edu.cn}

\affil[1]{\orgdiv{School of Instrument Science and Engineering}, \orgname{Southeast University}, \orgaddress{\city{Nanjing}, \postcode{210096}, \country{China}}}
\affil[2]{\orgdiv{The State Key Laboratory of Bioelectronics}, \orgname{Southeast University}, \orgaddress{\city{Nanjing}, \postcode{210096}, \country{China}}}

\abstract{\input{revision_abstract.tex}}
\maketitle

\noindent\textbf{ORCID:} Yuwen Li: \href{https://orcid.org/0000-0003-3060-836X}{0000-0003-3060-836X}\par

\section{Introduction}

Model comparison in EEG emotion recognition depends on how a trained model is selected as well as how it is fitted. Researchers commonly save a model after each training epoch and choose one checkpoint for evaluation. A higher score after searching more checkpoints can reflect useful differences between model states, favorable variation in the selection trials, or both. The practical issue is how much of the observed improvement persists when different trials score the selected model.

The need to separate model selection from final evaluation is established \citep{varma2006bias,cawley2010selection,kapoor2023leakage}. EEG studies have also documented the consequences of trial overlap and test-label use, while EEGain and LibEER provide substantial evaluation infrastructure \citep{brookshire2024leakage,lei2025leakage,kukhilava2025eegain,liu2025libeer}. These findings motivate a more specific comparison: along the same training trajectory, how do the score used for checkpoint selection and the score on other trials change as the candidate set expands? Reporting only the maximum cannot answer that question, and subtracting scores from different training protocols changes several factors at once.

We address this question through paired comparisons of saved checkpoints. Two disjoint trial pools exchange selection and scoring roles, while a separate validation pool and a fixed terminal checkpoint provide reference policies. A 300-trajectory same-session SEED experiment tests sensitivity to trial allocation; a further 276 trajectories apply a common cross-session design to SEED, SEED-IV and SEED-V. The latter comparison shows that part of the selected-score improvement can persist on other trials, qualifying the largely unretained gains in the same-session experiment. Complete curves, a common-time-range analysis and trial-level scores characterize the finding beyond a single selected maximum.

The contribution is an empirical account of retained checkpoint-selection gains under explicitly documented evaluation settings, supported by reusable reconstruction code. We distinguish \emph{same-session} evaluation on new trials from a known participant, \emph{cross-session} evaluation involving separate recordings from that participant, and \emph{cross-subject} evaluation on participants excluded from development. In the new cross-session experiment, target-session labels deliberately participate in checkpoint selection. This distinction is central to interpreting the retained gains.

\section{Related work}

SEED-family datasets supply repeated-session emotion recordings and differential-entropy representations widely used in EEG classification \citep{zheng2015seed,duan2013differential,zheng2019emotionmeter,liu2021seedv}. Graph-based models exploit spatial relationships between electrodes, but their reported accuracy remains conditional on the data partition and selection procedure \citep{song2020dgcnn,li2021sognn}. EEGain and LibEER standardize important parts of these procedures \citep{kukhilava2025eegain,liu2025libeer}; the present analysis uses a pinned LibEER DGCNN implementation rather than introducing another training library.

Lei et al. compared normal evaluation with test-selection, trial-wise and combined leakage conditions \citep{lei2025leakage}. Del Pup et al. compared sample-based, subject-based and nested validation, and Gil and Hern\'{a}ndez-Sabat\'{e} examined hierarchical evaluation units \citep{delpup2025partitioning,gil2026acrosssubject}. Our comparison builds on those distinctions by holding each training trajectory fixed and displaying four scoring policies over the candidate grid. Exchanging two complete trial pools makes it possible to quantify what remains of a selected-score improvement on the other pool.

Selection bias, clustered observations and uncertainty from overlapping fitted folds have a broader statistical literature \citep{cawley2010selection,bengio2004variance,bates2024cv,varoquaux2018cv}. We use these established principles to define the selection and scoring roles and to summarize paired participant records. The nonnegative difference between own-pool and exchanged-pool selection scores follows from maximization; its sign is not a new finding. The empirical questions concern its magnitude and the accompanying changes on other trials across the evaluated settings.

\section{Methods}

\subsection{Datasets and evaluation units}

The same-session experiment used official one-second SEED differential-entropy features smoothed with a linear dynamical system (DE/LDS), from 15 participants and sessions 1--2. The cross-session extension used all three sessions of SEED, SEED-IV and SEED-V. Table~\ref{tab:datasets} summarizes the released inputs. A complete film trial remained within one partition. No trial was divided between training, validation and target pools.

\begin{table}[!htbp]\centering\small
\caption{Released inputs for the cross-session extension. Counts are per dataset; participant overlap between datasets was not established.}\label{tab:datasets}
\begin{tabularx}{\textwidth}{@{}lccYc@{}}\toprule
Dataset & Participants & Classes & Released representation & Trials/session\\\midrule
SEED & 15 & 3 & One-second DE/LDS & 15\\
SEED-IV & 15 & 4 & Four-second DE/LDS & 24\\
SEED-V & 16 & 5 & Released DE, 310 features/window & 15\\
\bottomrule\end{tabularx}\end{table}

Each window was represented as 62 channels by five frequency bands. SEED-V vectors were reshaped accordingly, with trial labels read from the release; its official documentation identifies session numbers by stimulus set rather than temporal order \citep{seedvdocs}. The experiment therefore does not model prospective prediction of a chronologically later session. Structural checks covered 138 participant-session recordings, including finite features, class labels and complete trial mappings. The inputs are fixed released representations: additional standardization was fitted on training data only, but upstream smoothing and normalization were not independently reconstructed. In particular, the new SEED-IV four-second release differs from the historical one-second raw-to-feature pathway examined in the supplementary audit.

\subsection{Selection and scoring roles}

For every fitted trajectory, we retained 80 epochs and considered nested candidate counts $K\in\{1,5,10,20,40,80\}$. At count $K$, available epochs were $80/K,2(80/K),\ldots,80$. Checkpoints were selected by maximum window accuracy, with ties resolved at the earliest candidate epoch. The grid holds the trajectory and training horizon fixed while changing count, spacing and the earliest available epoch together.

Let $h_A$, $h_B$ and $h_V$ be checkpoints selected on trial pools A, B and validation, respectively, and $a_D(h)$ the window accuracy of checkpoint $h$ on pool $D$. We compared
\begin{align}
S_K &= \tfrac12\{a_A(h_A)+a_B(h_B)\},\\
C_K &= \tfrac12\{a_B(h_A)+a_A(h_B)\},\\
V_K &= \tfrac12\{a_A(h_V)+a_B(h_V)\}.
\end{align}
Here $S$ is the mean selection-pool score, $C$ is the mean score after exchanging selection and scoring pools, and $V$ uses the separate validation pool. A fourth policy always uses epoch 80. The symmetric mean weights the two pool roles equally. For the selection metric, $S_K-C_K\geq0$ by maximization. We therefore examine effect sizes and changes in $C$, rather than test whether this algebraic gap is positive. At $K=1$, all four policies coincide.

Both A and B are scored throughout the fixed trajectory and used retrospectively for their respective selection roles. These scores do not change gradient updates, normalization, training duration or model settings. For a given direction, the scoring pool does not select that checkpoint; nevertheless, the symmetric analysis uses labels from both pools and neither is a globally untouched test set. In the cross-session extension, both pools belong to the target session.

\begin{figure*}[!htbp]\centering
\includegraphics[width=0.98\textwidth]{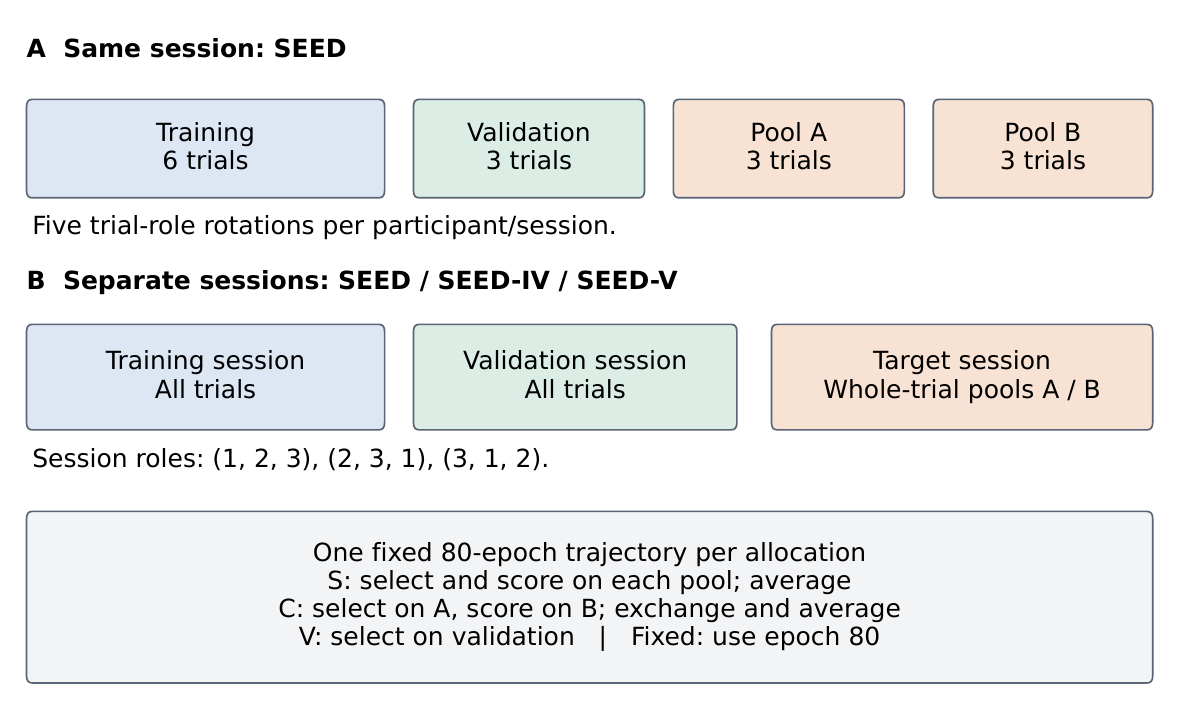}
\caption{Trial and session roles in the two designs. Complete trials stay within their assigned role. The schematic shows one session-role rotation; all three rotations are retained in the cross-session extension. Both A and B provide labels for their own checkpoint selection, and exchanging roles gives the cross-scored outcome. Validation selection and fixed epoch 80 provide separate reference policies.}\label{fig:design}
\end{figure*}

\subsection{Same-session design and common-time-range analysis}

Within each of the three SEED classes, five trials were permuted with seed 20260911 and cyclically assigned to training, training, validation, A and B. Across five rotations, each trial trained twice and served each other role once. Every trajectory used six training trials and three trials in each remaining pool. Fifteen participants, two sessions, two models and five rotations produced 300 trajectories with optimization seed 2024. DGCNN used the released DE/LDS inputs without additional standardization; the MLP used mean and population standard deviation fitted on its training trials, with standard deviations floored at $10^{-6}$. This normalization choice followed an earlier disclosed MLP fitting diagnostic (Online Resource 1, Section S9).

After examining the completed experiment, we computed paired changes from five to 80 candidates and added a common-time-range analysis. The latter uses the existing trajectories over epochs 16--80. Each candidate set includes both endpoints; 200 random permutations of the interior epochs generate nested sets of 5, 10, 20, 40 and 65 candidates (seed 20260913). The same sets are used across participants and models. This assesses whether the pattern depends on introducing epochs 1--15, while still allowing candidate position and correlation to change. No model was refitted. The plan was saved before this computation, after the earlier results were known.

\subsection{Cross-session design}

For each participant, training, validation and target sessions rotated through $(1,2,3)$, $(2,3,1)$ and $(3,1,2)$. All trials in the training and validation sessions served those roles. Within each target-session class, whole trials were randomly allocated to A and B, with A receiving the floor of half the class count and B receiving the remainder. Thus A/B contained 2/3 trials per class for SEED, 3/3 for SEED-IV and 1/2 for SEED-V. The split seed was 20260912 plus the target-session number, and the resulting trial IDs were shared across participants within each dataset/session. Class labels, split IDs and source checksums are recorded in the local input audit and frozen configuration.

Both models used per-feature mean and population standard deviation fitted solely on the training session, with the same $10^{-6}$ floor. The complete design comprised 90 SEED, 90 SEED-IV and 96 SEED-V trajectories. The six initial structural pilot cells remained in the matrix unchanged. All 276 trajectories were retained; low training accuracy was not an exclusion or tuning criterion. The protocol specified $S_{80}-S_5$ and $C_{80}-C_5$ as primary descriptive comparisons before new model outcomes. Other policy contrasts, complete curves and trial scoring were secondary. Earlier datasets had already been used in the research program, so this extension is not independent confirmation on previously untouched data.

\subsection{Models and training}

Both experiments used DGCNN from pinned LibEER revision \texttt{39dc27e} and an MLP with layers 310--128--64--$c$, where $c$ is the number of emotion classes. The MLP used ReLU activations and dropout 0.5 after each hidden layer. DGCNN retained the pinned implementation and its 0.01 norm penalty; source and configuration checksums are supplied with the code. Neither architecture was tuned to the new outcomes.

Training used 80 epochs, batches of 64, cross-entropy and AdamW with learning rate 0.0015, weight decay $10^{-4}$ and epsilon $10^{-4}$. Optimization seed 2024 was reset for each trajectory. Evaluation used deterministic batches of 512, with dropout disabled. The cross-session extension ran with deterministic PyTorch settings on an NVIDIA RTX 4060. Saved arrays include per-epoch window scores, per-epoch trial logits, and final-policy window predictions. The same-session and cross-session experiments differ in training size, trial-role allocation and DGCNN standardization; their contrast cannot isolate a causal session effect.

\subsection{Statistical summaries and reconstruction}

Window accuracy was the selection metric and primary scoring endpoint. For the secondary trial endpoint, logits were averaged within each complete trial before taking the most probable class; trial accuracy then weighted trials equally within each pool. Checkpoints were not reselected for trial scoring, so the algebraic nonnegativity of the window-score gap need not hold for trial scores.

We first averaged dependent sessions and rotations within each participant. In the common-time-range analysis, candidate-set repetitions were also averaged within participant. Paired percentile bootstrap intervals used 20,000 resamples of these participant records: seed 20260911 for the same-session experiment, 20260913 for the common-time-range analysis and 20260912 for the cross-session extension. Policy differences were calculated within participant before resampling. We report descriptive, unadjusted 95\% intervals; an interval crossing zero does not establish equivalence, and different interval-crossing outcomes do not establish a between-condition difference.

The statistical units remain 15 participants for SEED, 15 for SEED-IV and 16 for SEED-V, analyzed separately. Sessions, rotations and candidate sets do not increase these counts. Intervals summarize participant-record variability conditional on the empirical stimuli, partitions and realized fits; they do not cover full retraining variability or a new population by construction. A separate implementation reconstructed saved metrics and paired summaries. Synthetic examples also check label consistency, trial aggregation, selection ties, partition overlap and declared event order without requiring restricted EEG data (Online Resource 1, Sections S11--S14).

\paragraph{Computational assistance.} OpenAI Codex assisted with code drafting, execution orchestration, numerical reconstruction and manuscript preparation. Its role and the verification of computational outputs are disclosed in Statements and Declarations.

\section{Results}

\subsection{Cross-session gains persisted on other trials to differing degrees}

All 276 planned trajectories completed. Mean terminal training-window accuracy was 100\% in five of the six dataset/model combinations. SEED-IV DGCNN averaged 99.58\%, with a minimum of 81.20\% across its 45 trajectories; this lower-fit trajectory was retained. These diagnostics show that simple failure to fit the training data was not a general explanation for the observed patterns, without excluding other implementation or distributional factors.

Increasing the candidate set raised selection-pool accuracy for both models on all three datasets. For DGCNN, part of that improvement persisted after selection and scoring pools were exchanged: all three descriptive intervals for $C_{80}-C_5$ were positive (Table~\ref{tab:family}). MLP retained a positive change on SEED, whereas its SEED-IV and SEED-V intervals included zero. Thus the experiment does not support a blanket claim that gains from additional checkpoint selection disappear on other trials.

\begin{table*}[!htbp]\centering\small
\caption{Cross-session changes from five to 80 candidates. Values are percentage points with descriptive 95\% paired participant-bootstrap intervals; $n=15$, 15 and 16 for SEED, SEED-IV and SEED-V. Both target pools supply labels for symmetric selection.}\label{tab:family}
\begin{tabular}{@{}llcc@{}}\toprule
Dataset & Model & $S_{80}-S_5$ & $C_{80}-C_5$\\\midrule
SEED & DGCNN & 8.06 [6.38, 10.02] & 4.47 [1.79, 7.25]\\
SEED & MLP & 5.14 [2.87, 7.81] & 4.20 [1.81, 7.16]\\
SEED-IV & DGCNN & 7.95 [6.07, 10.07] & 4.27 [2.00, 6.53]\\
SEED-IV & MLP & 6.21 [4.79, 7.88] & 2.20 [-0.18, 4.89]\\
SEED-V & DGCNN & 9.08 [7.30, 10.92] & 2.75 [0.79, 4.78]\\
SEED-V & MLP & 4.67 [3.04, 6.42] & 0.23 [-1.77, 2.16]\\
\bottomrule\end{tabular}\end{table*}

\begin{figure*}[!htbp]\centering
\includegraphics[width=0.98\textwidth]{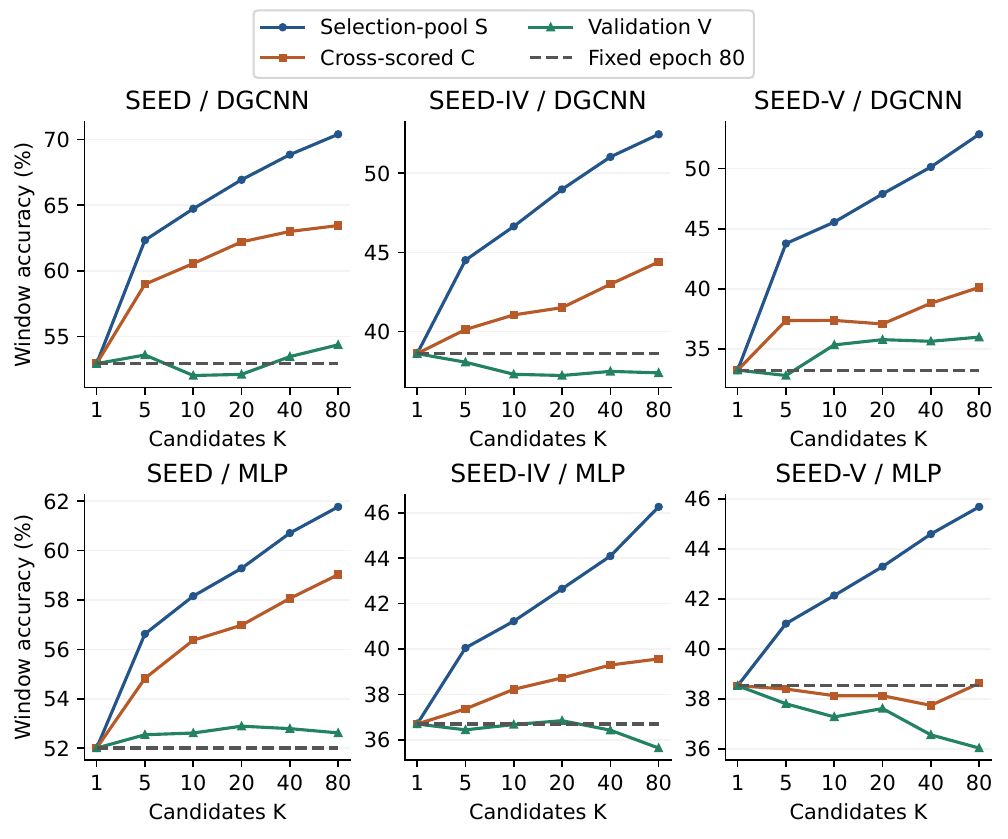}
\caption{Complete cross-session policy curves. Each panel shows participant-equal mean window accuracy after averaging three session-role rotations within participant. Candidate counts occupy equally spaced categorical positions. Panel-specific vertical scales aid inspection within each dataset/model; absolute scores across tasks are not a common leaderboard. Fixed epoch 80 is shared by all candidate sets.}\label{fig:family}
\end{figure*}

The validation-selected and fixed-epoch curves provide a different reference (Fig.~\ref{fig:family}). Changes in validation-selected accuracy from five to 80 candidates had intervals including zero in all six combinations (Online Resource 1, Section S14). The positive cross-scored changes therefore concern checkpoints chosen using target-session labels; they do not demonstrate improved performance without target-label access. Complete-trial scoring retained the DGCNN direction on all three datasets and the same pattern of interval crossing for MLP. These secondary scores come from the same participants and models, not an independent replication.

\subsection{Same-session gains were largely confined to the selection pools}

The 300-trajectory SEED same-session experiment showed a different pattern. DGCNN selection-pool accuracy increased by 6.24 percentage points [5.24, 7.33] from five to 80 candidates, whereas the cross-scored change was $-1.24$ [$-2.67$, 0.26]. The corresponding MLP changes were 1.96 [1.48, 2.49] and $-0.16$ [$-0.66$, 0.39] points. Both models reached terminal training accuracy 1.000. At 80 candidates, cross-scored differences from validation selection were 0.03 [$-1.61$, 1.60] points for DGCNN and $-0.13$ [$-0.64$, 0.36] for MLP. These intervals do not establish equivalence or deterioration.

\begin{figure*}[!htbp]\centering
\includegraphics[width=0.98\textwidth]{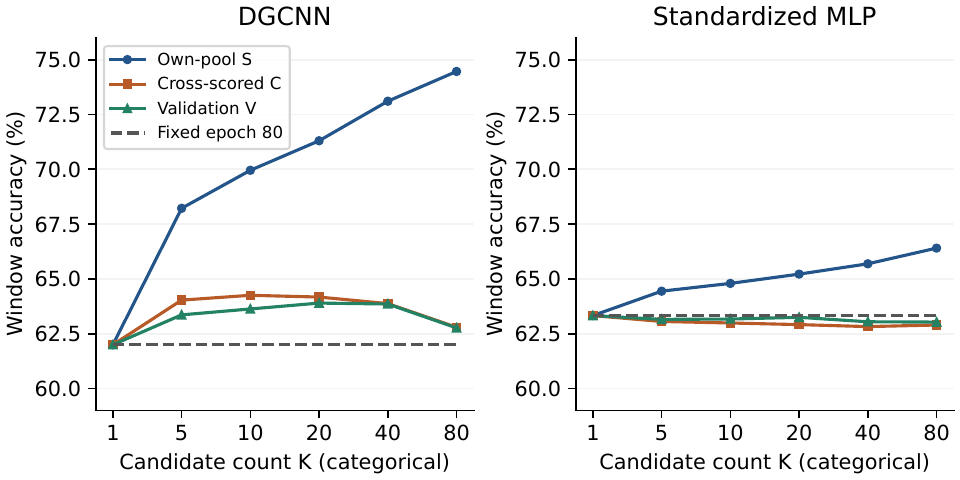}
\caption{Same-session SEED policy curves. Means average two sessions and five dependent trial-role rotations within each of 15 participants. DGCNN uses released features without additional standardization; MLP uses training-only standardization. Candidate count, spacing and earliest available epoch change jointly. Selected-epoch distributions and all means are provided in Online Resource 1, Section S12.}\label{fig:samesession}
\end{figure*}

Restricting the available time range to epochs 16--80 preserved the separation. Across 200 nested candidate-set repetitions, DGCNN selection-pool accuracy increased by 4.58 [4.10, 5.10] points from five to 65 candidates, while cross-scored accuracy changed by $-0.07$ [$-0.78$, 0.66]. MLP changes were 0.69 [0.54, 0.83] and $-0.09$ [$-0.28$, 0.09] points. The same-session pattern therefore did not require the introduction of epochs 1--15. Candidate positions and correlation still varied (Online Resource 1, Section S13).

\subsection{Reconstruction and historical context}

Separate local implementations reproduced the saved cross-session scores and participant summaries and the common-time-range analysis. Cross-session checks covered source and output checksums, train-only standardization, all-epoch trial scores, final-policy window predictions and the reported paired intervals. Other budget-specific window checks used saved scalar accuracies. Eleven synthetic cases tested the reusable audit utility; these demonstrate specified arithmetic and consistency checks, not a real-world detection rate for undocumented data use.

Historical results explain the motivation and preserve the study's development record (Online Resource 1, Sections S1--S12 and S15). The archived SEED best-test versus epoch-80 difference was 10.36 percentage points, but it alone cannot identify a pure leakage penalty. The archived SEED implementation met its public-reference tolerance; SEED-IV remained 3.40 points below its reference after the documented audit. That unresolved one-second-pipeline comparison does not certify or invalidate the distinct four-second released-feature extension. The old strict cross-subject endpoints remain separate supporting analyses. Likewise, unstable participant rankings and the lack of a consistent tail-mixture accuracy advantage constrain the original subgroup and algorithm claims; those outcomes are retained in the supplement rather than treated as evidence for the new checkpoint comparison.

\section{Discussion}

\subsection{Selected-score improvements and retained gains answer different questions}

The experiments show why a checkpoint-selection result should include scores on trials other than those selecting the checkpoint. In the same-session SEED design, a larger candidate set primarily raised the selection-pool score. Under the new cross-session design, DGCNN retained part of the increase on other target-session trials across all three datasets, while MLP results were less consistent. A selected maximum alone would conceal this distinction. Conversely, treating all of its increase as spurious would overlook the retained gains.

Useful model-state variation and selection on trial-specific variation can coexist along a training trajectory. The present measurements quantify the resulting scores without decomposing those mechanisms. In particular, differences between the same-session and cross-session experiments also involve training-set size, pool composition and normalization. They do not identify session change as the cause. Nor do retained gains justify reusing a final evaluation set for an independent performance claim: the target labels are an explicit part of the selection procedure here.

\subsection{Contribution to EEG evaluation practice}

Prior work establishes the risks of test selection and inappropriate grouping, and provides broad EEG benchmarking infrastructure \citep{lei2025leakage,kukhilava2025eegain,liu2025libeer}. This study adds paired budget curves showing both the selected score and the accuracy retained on other trials, with role exchange, a common-time-range check and a three-dataset extension. It makes the amount retained an observed quantity rather than assuming either complete transfer or complete disappearance. The reconstruction code connects those quantities to the actual candidate epochs and scoring rules.

Three reporting practices follow. First, identify the data used to select the checkpoint, including any target-session labels. Second, report the candidate epochs and a separate scoring comparison alongside the maximum; a fixed endpoint and validation-selected policy clarify the reference. Third, state the prediction unit and participant-level aggregation so that many windows or repeated roles are not mistaken for many independent participants. Table~\ref{tab:reporting} gives a compact record that can accompany existing training libraries.

\begin{table}[!htbp]\centering\small
\caption{Minimum record for interpreting a checkpoint-selection comparison.}\label{tab:reporting}
\begin{tabularx}{\textwidth}{@{}lY@{}}\toprule
Item & Information to retain\\\midrule
Evaluation target & Dataset/feature release; participant and session relation; complete-trial partition\\
Model selection & Candidate epochs; selection metric and tie rule; validation and target-label access\\
Reference policies & Fixed checkpoint and separate validation rule, evaluated on the same scored trials\\
Scoring & Window or trial endpoint; trial aggregation; participant weighting\\
Uncertainty & Paired differences; resampling unit; dependent rotations; descriptive or confirmatory status\\
Reconstruction & Predictions or sufficient scores; split IDs; configuration and code version; prior data use\\
\bottomrule\end{tabularx}\end{table}

\subsection{Scope and limitations}

The three releases broaden task coverage within the SEED family, but are not three established independent populations. Participant overlap between datasets is unknown, all datasets were previously encountered, and the samples contain 15--16 participants each. Each extension was recorded before its own execution after earlier results had been inspected. The cross-session comparison uses one optimization seed and three dependent rotations; its intervals do not quantify complete retraining or recruitment uncertainty. The common-time-range analysis removes one temporal-coverage difference without isolating a pure candidate-count effect.

The conclusions begin at the released feature representation. We verified the additional training-only standardization, not the independence of every upstream feature operation. Target-session trials share participants and recording conditions, and the unequal A/B sizes in SEED and SEED-V differ from the balanced same-session pools. Model-specific and trial-specific responses may therefore contribute to retained gains. These are protocol-specific measurements, not estimates of performance on new participants or a universal correction for reported accuracy.

The pinned DGCNN implementation, its historical SEED-IV discrepancy and the original negative analyses remain documented in the supplement. High training fit excludes only a simple inability to fit the observed training data. The audit utility verifies supplied arrays and logs, not omitted researcher actions. A future study seeking a causal explanation of the between-design differences would need to hold representation, standardization and training allocation constant while changing the session relation.

\section{Conclusion}

Checkpoint search should be evaluated by what its selected-score improvement retains on other trials. The SEED-family experiments exhibited both largely unretained same-session gains and partially retained gains when target-session labels selected among models trained on another session. Reporting the selection rule, complete candidate grid and separate scoring outcome makes that distinction visible. The practical result is a reproducible comparison that can accompany EEG benchmark scores without treating a selected maximum as an independent generalization estimate.

\section*{Statements and Declarations}

\subsection*{Information Sharing Statement}

SEED, SEED-IV and SEED-V are available to eligible academic researchers through the official application portal under the source-data licence. The competition release is available through the organizer subject to its terms. The archived analysis repository is \url{https://github.com/hantingsuo/eeg-generalization-research}. The version v4-2026-09-12 local reviewer package, Online Resource 3 (\texttt{Reproducibility\_code.zip}), contains versioned revision code, frozen protocols, environment instructions, synthetic correctness examples and aggregate figure inputs. It contains no raw EEG, participant-level prediction arrays or model checkpoints. Synthetic examples and aggregate plots can be run without the source data; full training requires authorized source data. The versioned revision package is supplied as Online Resource 3 for review and distribution with the article. Original revision code is licensed under BSD-3-Clause; the included LibEER files retain their MIT licence. Online Resources 1 and 2 retain supporting analyses and full-precision mixture configurations.

\subsection*{Acknowledgements}

The authors thank the organizing committee of the 11th National College Student Biomedical Engineering Innovation Design Competition and the Task 4 team of its Brain-Computer Interface Track, including Pazhou Laboratory, for providing the competition data, documentation, and evaluation platform.

\subsection*{Funding}

This work was supported in part by the National Natural Science Foundation of China under Grant 62571123; in part by the Basic Research Program of Jiangsu Province under Grant BK20252010; in part by the Fundamental Research Funds for the Central Universities (2242026RCB0024).

\subsection*{Competing interests}

The authors have no relevant financial or non-financial interests to disclose.

\subsection*{Author contributions}

Hanting Suo: Conceptualization, Methodology, Software, Validation, Formal analysis, Investigation, Data curation, Visualization, Writing -- original draft, Writing -- review and editing. Yuwen Li: Conceptualization, Methodology, Supervision, Project administration, Resources, Funding acquisition, Writing -- review and editing. Hongxun Wang: Data curation, Software.

\subsection*{Related preprint}

An earlier version of this work is available as arXiv:2607.27655 (\url{https://arxiv.org/abs/2607.27655}), under the title \emph{Evaluation Protocols and Cross-Subject Generalization in EEG Emotion Recognition}. The present manuscript has a revised focus, additional experiments and an updated author list.

\subsection*{AI-assisted work}

OpenAI Codex assisted with revision planning, code drafting, execution orchestration, numerical reconstruction, simulated manuscript critique and English manuscript editing. The simulated critiques are internal checks, not journal peer review. The authors retain responsibility for the scientific claims, references and submitted materials. Computational outputs were checked through separate numerical reconstruction and runnable synthetic examples.

\subsection*{Ethics approval and consent to participate}

Not applicable to the present secondary analysis. The authors did not recruit or contact participants, conduct interventions, or access directly identifying information. Ethical oversight and consent for the original data collection remain the responsibility of the respective data providers. The datasets were used in accordance with their applicable access and licence terms.

\subsection*{Consent for publication}

Not applicable. This article contains no identifiable participant information, images, or case details.

\subsection*{Data availability}

The SEED, SEED-IV and SEED-V data were obtained from the BCMI laboratory under the SJTU Emotion EEG Dataset License Agreement and are available to eligible academic researchers through the \href{https://bcmi.sjtu.edu.cn/ApplicationForm/apply_form/}{official application portal}. The licence permits academic research use but prohibits redistribution; raw SEED, SEED-IV and SEED-V files are therefore not included with this article. The competition release can be requested or downloaded through the organizer's \href{https://www.pazhoulab.com/2026/03/8165/}{official Task 4 page} subject to the current access and reuse terms. The authors do not redistribute the competition data.

\subsection*{Code and material availability}

Online Resource 1 (\texttt{Supplementary\_information.pdf}) contains the complete secondary checkpoint results, common-time-range analysis, historical implementation and training-fit checks, and the original supporting analyses. Online Resource 2 (\texttt{ESM\_2.json}) provides the exact CF-TRE component order, selected mean- and tail-risk parameters, double-precision mixture weights, and source-file checksum in machine-readable form. The archived pre-revision analysis code and configurations are available at \href{https://github.com/hantingsuo/eeg-generalization-research}{https://github.com/hantingsuo/eeg-generalization-research}. The v4-2026-09-12 revision code, aggregate inputs, synthetic examples and versioned manifest clarification are supplied in Online Resource 3 (\texttt{Reproducibility\_code.zip}); it is supplied for review and distribution with the article under BSD-3-Clause for original code and the retained upstream licence for LibEER. Raw SEED, SEED-IV, SEED-V, and competition data, together with fine-grained participant-level predictions governed by the source-data terms, are not redistributed.

\begingroup
\widowpenalty=10000
\clubpenalty=10000
\interlinepenalty=10000
\emergencystretch=3em
\fontsize{8.8pt}{10.6pt}\selectfont
\setlength{\bibsep}{0.35em}
\bibliography{references}
\endgroup
\end{document}

%% file: revision_abstract.tex
Checkpoint selection can improve an electroencephalography (EEG) emotion-recognition score without improving performance on other trials. We compared selection and scoring on disjoint trial pools along fixed training trajectories. A same-session SEED study comprised 300 trajectories from 15 participants, two models and five trial-role rotations. Another 276 trajectories extended the comparison to separate training, validation and target sessions in SEED, SEED-IV and SEED-V, with 15, 15 and 16 participants, respectively. Increasing the candidate set from five to 80 checkpoints raised the same-session dynamical graph convolutional neural network (DGCNN) selection-pool score by 6.24 percentage points, while the other-pool change was $-1.24$ (descriptive 95\% participant-bootstrap interval $-2.67$ to 0.26). Under the cross-session design, DGCNN retained gains of 4.47 [1.79, 7.25], 4.27 [2.00, 6.53] and 2.75 [0.79, 4.78] points across the three datasets. Multilayer perceptron results varied across datasets. Common-time-range analysis and trial-level scoring supported the respective patterns. Both target pools supplied labels for symmetric checkpoint selection; these results concern retention across trials, not evaluation without target labels. Complete policy curves and executable reconstruction support reporting selected-score improvements alongside performance on other trials, rather than assuming a universal penalty for checkpoint search.